\documentclass[letterpaper]{article} 
\usepackage{aaai24}  
\usepackage{times}  
\usepackage{helvet}  
\usepackage{courier}  
\usepackage[hyphens]{url}  
\usepackage{graphicx} 
\usepackage{natbib}  
\usepackage{caption} 
\usepackage{adjustbox}
\usepackage{algorithm}
\usepackage{algorithmic}
\usepackage{amsmath}
\usepackage{amssymb}
\usepackage{array}
\usepackage{arydshln}
\usepackage{booktabs}
\usepackage{enumitem}
\usepackage{makecell}
\usepackage{multirow}
\usepackage{multicol}
\usepackage{utfsym}
\usepackage{subcaption}

\usepackage{newfloat}
\usepackage{listings}
\DeclareCaptionStyle{ruled}{labelfont=normalfont,labelsep=colon,strut=off} 
\floatstyle{ruled}
\newfloat{listing}{tb}{lst}{}
\floatname{listing}{Listing}
\DeclareMathOperator*{\argmax}{arg\,max}

\title{Collaborative Weakly Supervised Video Correlation Learning \\ for Procedure-Aware Instructional Video Analysis}

\author{
    Tianyao He\textsuperscript{\rm 1},
    Huabin Liu\textsuperscript{\rm 1},
    Yuxi Li\textsuperscript{\rm 1},
    Xiao Ma\textsuperscript{\rm 2},
    Cheng Zhong\textsuperscript{\rm 2},
    Yang Zhang\textsuperscript{\rm 2},
    Weiyao Lin\textsuperscript{\rm 1}\thanks{Corresponding author.}
}
\affiliations{
    \textsuperscript{\rm 1}Shanghai Jiao Tong University, Shanghai, China \\
    \textsuperscript{\rm 2}AI Lab, Lenovo Research, Beijing, China \\
    \{hetianyao,huabinliu,lyxok1\}@sjtu.edu.cn,\\
    \{maxiao3, zhongcheng3, zhangyang20\}@lenovo.com,
    wylin@sjtu.edu.cn
}

\begin{document}

\maketitle

\begin{abstract}
    Video Correlation Learning (VCL), which aims to analyze the relationships between videos, has been widely studied and applied in various general video tasks. However, applying VCL to instructional videos is still quite challenging due to their intrinsic procedural temporal structure. Specifically, procedural knowledge is critical for accurate correlation analyses on instructional videos. Nevertheless, current procedure-learning methods heavily rely on step-level annotations, which are costly and not scalable. To address this problem, we introduce a weakly supervised framework called Collaborative Procedure Alignment (CPA) for procedure-aware correlation learning on instructional videos. Our framework comprises two core modules: collaborative step mining and frame-to-step alignment. The collaborative step mining module enables simultaneous and consistent step segmentation for paired videos, leveraging the semantic and temporal similarity between frames. Based on the identified steps, the frame-to-step alignment module performs alignment between the frames and steps across videos. The alignment result serves as a measurement of the correlation distance between two videos. We instantiate our framework in two distinct instructional video tasks: sequence verification and action quality assessment. Extensive experiments validate the effectiveness of our approach in providing accurate and interpretable correlation analyses for instructional videos.
\end{abstract}

\begin{figure}[t]
    \centering
    \includegraphics[width=1.0\linewidth]{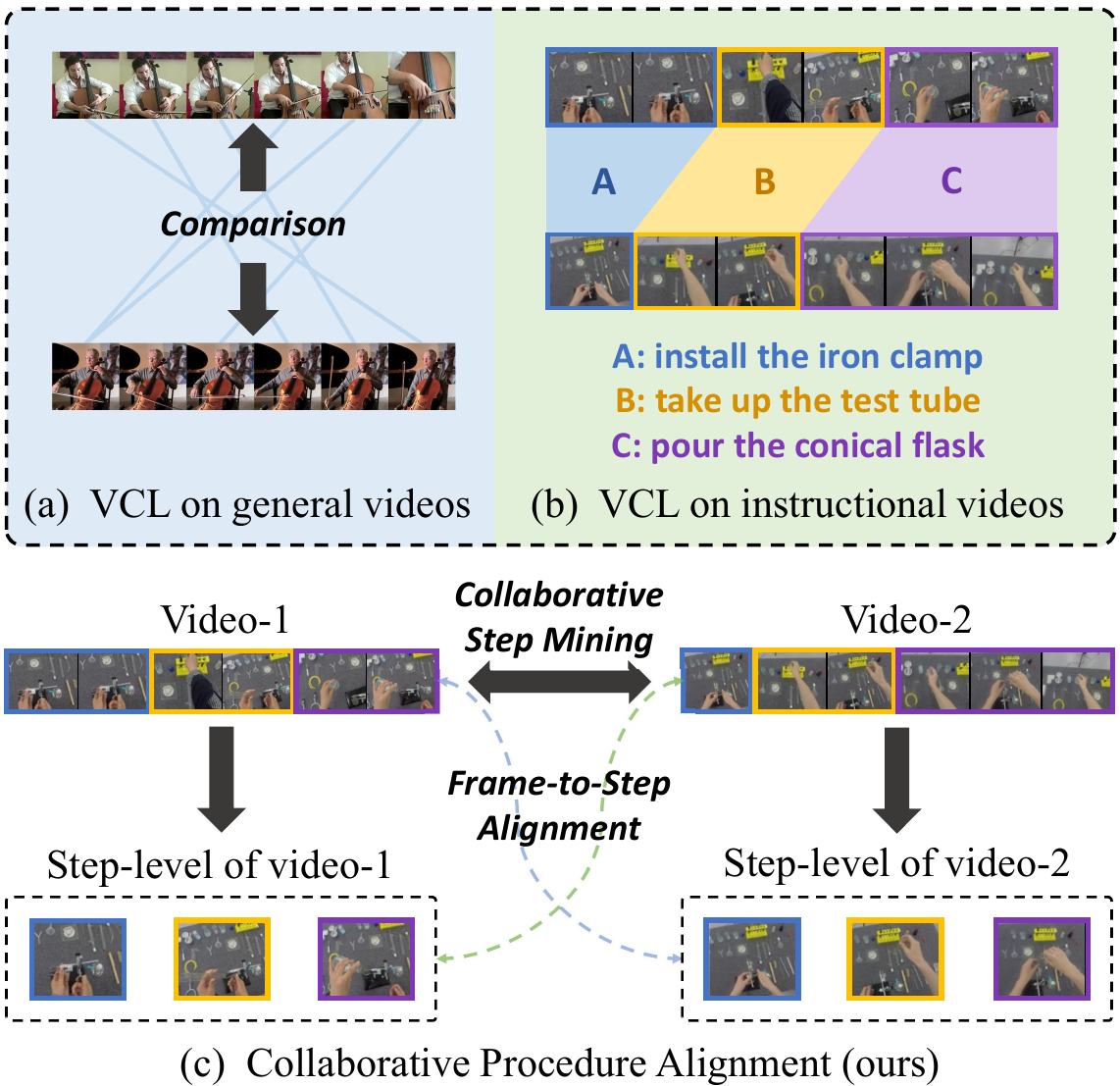}
    \caption{(a)(b) VCL on general videos and instructional videos; (c) Our Collaborative Procedure Alignment framework, which conducts a procedural alignment between frames and collaboratively mined steps.}
    \label{fig:head}
\end{figure}

\section{Introduction}
    Video Correlation Learning (VCL) focuses on examining and quantifying the relationships between videos through a comparative paradigm. It empowers researchers to discover temporal and conceptual knowledge from the intrinsic associations between videos. Numerous previous studies have explored VCL in general videos. For example, some methods adopt VCL to grasp the similarities and differences between videos, including video contrastive learning~\cite{qian2021spatiotemporal, park2022probabilistic}, and video retrieval~\cite{zhou2018temporal, wang2019learning}. Other studies utilize VCL to analyze individual videos referring to given exemplars (or called support videos), such as video quality assessment~\cite{mozhaeva2021full, xu2020c3dvqa}, and few-shot action recognition~\cite{zhu2018compound, ben2021taen}. As depicted in Fig.~\ref{fig:head}(a), VCL in general videos primarily centers on the video and frame-level comparison.
    
    However, applying traditional VCL approaches to \textit{instructional} videos encounters significant challenges. Specifically, instructional videos comprise multiple fine-grained steps with varying durations and temporal locations. This results in more complex procedural structures compared to general videos (see Fig.~\ref{fig:head}(b)). Therefore, to achieve precise and interpretable correlation learning for instructional videos, the crux lies in capturing procedural knowledge.
    
    Currently, many procedure-learning methods for instructional videos are emerging~\cite{behrmann2022unified,han2022temporal,xu2022finediving}. However, they heavily rely on step-level annotations. These annotations require step semantic labels and their temporal boundaries, incurring substantial costs and lacking scalability. This naturally raises a pivotal question: \textit{How can we learn the intrinsic procedural knowledge of instructional videos without step-level annotations?} In this paper, we present an insight into this question for VCL: \textit{Two instructional videos that share the same procedure often exhibit a strong internal correlation between steps, which can serve as a valuable reference for mutual procedure learning.}

    Based on this insight, we present a weakly supervised collaborative procedure alignment (CPA) framework to achieve procedure-aware correlation learning for instructional videos. Here, ``weakly supervised'' refers to accessing only video-level classes while step-level annotations are unknown. Videos belonging to the same video-level class present identical procedures. As illustrated in Fig.~\ref{fig:head}(b), our framework harnesses the internal correlation between paired videos, allowing for the collaborative extraction of step-level information and quantifying the step-level correlation between instructional videos. Specifically, our framework consists of two core components: collaborative step mining and frame-to-step alignment. Collaborative step mining exploits both semantic and temporal similarities among video frames, enabling the simultaneous extraction of steps for paired instructional videos. It empowers us to extract a video's steps with guidance from the other, and vice versa. Built upon the step-level features, we then design a frame-to-step alignment module to quantify the procedure consistency between the two videos. The alignment is performed between the step-level features of one video and the frame-level features of the other.  A higher alignment probability signifies a higher likelihood of step-level consistency. This probability serves as a distance quantifying the procedure correlation between these two instructional videos.

    We validate our framework by performing video correlation learning on two instructional video tasks, including \emph{sequence verification} and \emph{action quality assessment}. Extensive experiments show the effectiveness of our framework in providing more precise and interpretable predictions of correlation.

    Overall, our contributions can be summarized as follows:
    \begin{itemize}[leftmargin=*]
        \item We propose a weakly supervised collaborative procedure alignment framework for instructional video correlation learning, which collaboratively extracts consistent steps in paired videos and then measures their distance through a procedure alignment process.
        \item Under this framework, we devise a collaborative step mining approach accounting for the semantic and temporal relationships between videos, which enables concurrent step segmentation in paired videos. In addition, we introduce a frame-to-step alignment module to furnish a precise measure of video distance.
        \item We apply our framework to two instructional video tasks, including sequence verification and action quality assessment. Extensive experiments showcase the superiority of our framework, demonstrating its capacity to deliver more accurate and explainable results over existing competitors.
    \end{itemize}

\section{Related Work}
\subsubsection{Video correlation learning}
    Video correlation learning is a technology adopted by a wide range of work, which can be roughly divided into two streams. The first stream of work aims to learn the similarities and differences between two videos based on the given criteria. For example, in video retrieval, the method should find videos highly related to the query video. \cite{han2020self,zhang2020large} solve the task through video representation learning by comparing the video-level features. \cite{jo2023vvs, jo2023simultaneous} adopt frame-level and temporal information for more accurate predictions. Another stream analyzes the query video with reference to the exemplar videos. For example, few-shot action recognition aims to classify the query video based on only a few support videos. Some studies~\cite{cao2020few,hadji2021representation} adopt temporal alignment between videos, while other works have explored more flexible alignment strategies based on attention mechanisms~\cite{li2022ta2n,liu2022task,liu2023few}, and distribution distance~\cite{wu2022motion,wang2022hybrid}. Another example is video quality assessment. \cite{mozhaeva2021full,xu2020c3dvqa} assess the query video based on an exemplar whose quality score is given. Currently, the majority of the exploration on VCL focuses on general videos, while studies on instructional videos remain inadequate.

\subsubsection{Instructional video learning}
    Instructional videos are created to convey skills, knowledge, or procedural information, which find extensive usage in education, training, and demonstrations. Therefore, tasks related to instructional videos are gaining increasing attention. Related datasets including COIN~\cite{tang2019coin}, Diving~\cite{li2018resound}, CSV~\cite{qian2022svip}, EPIC-KITCHENS~\cite{damen2018scaling}, Assembly101~\cite{sener2022assembly101}, and HiEve~\cite{lin2023hieve} have provided instructional videos in different scenarios. A prominent task for instructional videos is action segmentation, which aims to divide a video into successive steps. In this field, \cite{richard2016temporal,singh2016multi, lea2017temporal,lei2018temporal,farha2019ms} necessitates step boundary annotations for segmentation, while \cite{aakur2019perceptual,sarfraz2021temporally,du2022fast,wang2022sscap} achieves unsupervised segmentation based on the semantic similarity and temporal continuity of frames within a step. Recently, novel instructional video tasks have been developed.  \cite{qian2022svip} proposed the sequence verification task, aiming at verifying whether two instructional videos have the same procedure. Additionally, \cite{xu2022finediving} proposed the procedure-aware action quality assessment task to score the diving sports videos based on a standard exemplar video. This paper focuses on correlation learning for instructional videos without relying on step-level annotations, which can be applied to various specific tasks. 
    
\begin{figure*}
    \centering
    \includegraphics[width=0.95\linewidth]{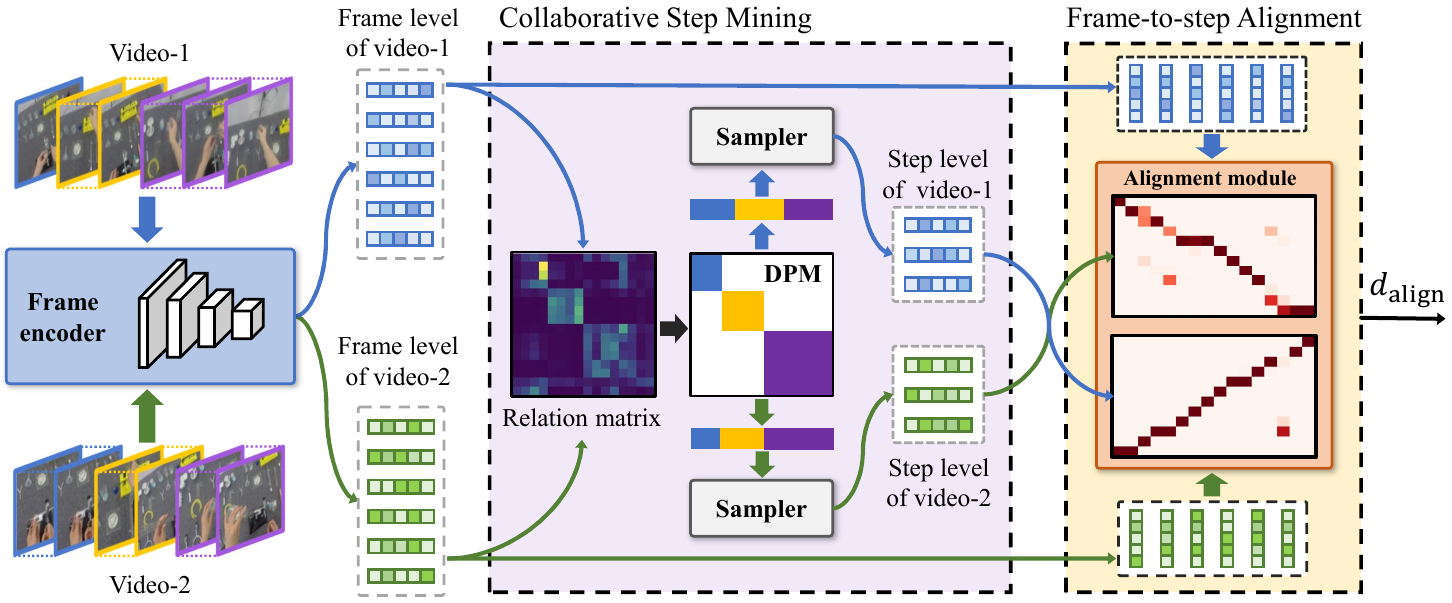}
    \caption{The pipeline of CPA. Based on the frame-level features, we utilize the collaborative step mining module (CSM) to produce the step segmentation of two videos simultaneously. Then, we can sample step-level features according to the step boundaries. Finally, we design a frame-to-step alignment (FSA) between two videos to get their correlation distance.}
    \label{fig:struct}
\end{figure*}

\section{Method}
\subsection{Overview}
\label{sec:cross-level}
    The overall pipeline of our CPA framework is illustrated in Fig.~\ref{fig:struct}. The input paired sample is $\{X_1, X_2; Y_1, Y_2\}$, where $X_1,X_2$ are two videos' frame/clip sets, and $Y_1,Y_2$ are their video-level labels. It is important to note that \textit{we do not use any step-level annotation under this setting.} To begin with, the video data are fed into the frame encoder $\Psi(\cdot)$ to generate frame-level features: $F_1=\Psi(X_1)=\{f_{1}^1, f_{2}^1, \dots, f_{T}^1\}$ and $F_2=\Psi(X_2)=\{f_{1}^2, f_{2}^2, \dots, f_{T}^2\}$. 
    Based on the frame-level features, we initially employ our collaborative step mining module to obtain a coherent step segmentation of the paired videos. For each step segment, we can sample corresponding step-level representation from frame features. Then, we apply the frame-to-step alignment between one video's frame-level features and another video's step-level features to produce the video distance.
    
\subsection{Collaborative Step Mining}
\begin{figure}[t]
        \centering
        \includegraphics[width=\linewidth]{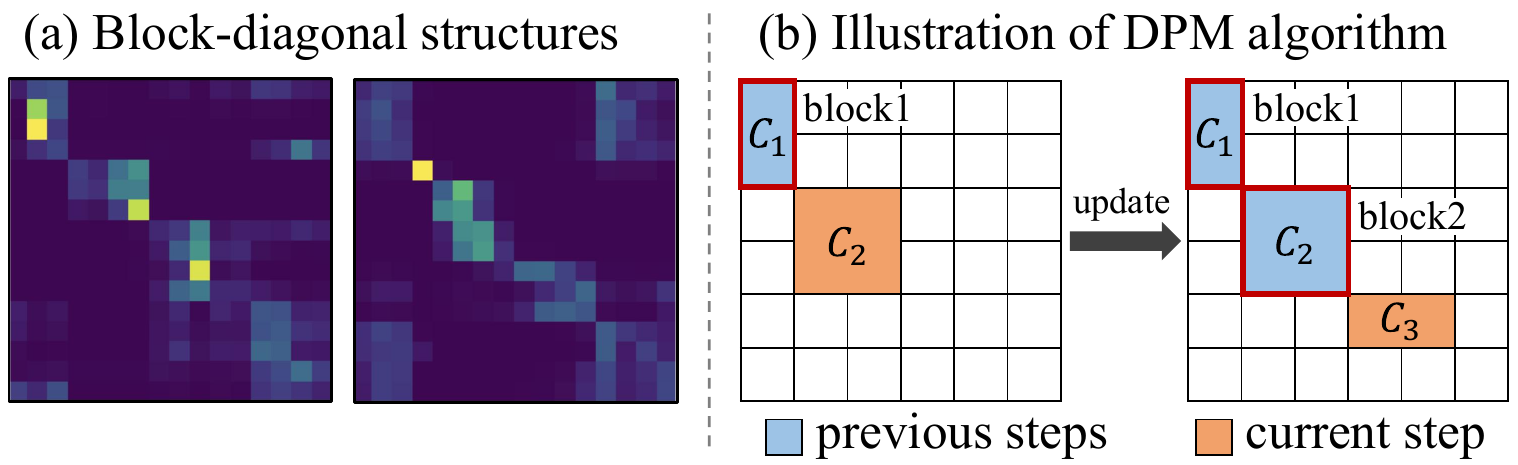}
        \caption{The illustration of: (a) the block-diagonal structure; (b) the DPM algorithm.}
        \label{fig:csm_method}
    \end{figure}

    In instructional videos, we have the observation~\cite{sarfraz2021temporally, du2022fast} that frames within the same step should have: (1) high semantic similarity and (2) continuous temporal order. Therefore, for two videos sharing the same procedure, their corresponding steps should also exhibit high semantic similarity and temporal continuity. As shown in Fig.~\ref{fig:csm_method}(a), we can observe block-diagonal structures in the relational matrix of two consistent instructional videos, where each block represents a coherent step segment. Consequently, we can achieve a consistent step segmentation for paired videos based on the block-diagonal structure. We propose a dynamic programming-based Collaborative Step Mining (CSM) module to extract the block-diagonal structure from the relational matrix so that we can produce step segments for paired videos simultaneously. Collaboratively extracting steps from paired videos can ensure the consistency of the step-level information from two videos, which takes advantage of their internal correlation. In this section, we will elaborate on how it works.

\subsubsection{Relational matrix calculation}
    First, we calculate two videos' relational matrix $\mathcal{M}$ by their frame-wise similarity: $\mathcal{M}=\mathrm{Softmax}\left[(F_1\cdot F_2^\mathrm{T}) / \sqrt{d}\right]$, where $d$ denotes their feature dimension number. $\mathcal{M}_{ij}$ means the similarity between the $i_{\mathrm{th}}$ frames of video-1 and the $j_{\mathrm{th}}$ frame of video-2.

\subsubsection{Dynamic procedure matching}
    Then, we design the dynamic procedure matching (DPM) to seek the best step segmentation between two videos by adopting the idea of dynamic programming. Since it remains uncertain how many steps (blocks) should be divided, we set the step number to $K$, which indicates we expect to partition the relational matrix into $K$ blocks (i.e., $K$ steps). We define the $K$ blocks as: $\mathcal{B}=\{\mathrm{block}_1, \mathrm{block}_2, \dots, \mathrm{block}_K\}$. Here, $\mathrm{block}_i=(a_i, b_i, x_i, y_i)$ where $(a_i, b_i)$ and $(x_i, y_i)$ are the top-left and bottom-right coordinates of $\mathrm{block}_i$ on the relational matrix. Consequently, we can calculate a consistency score for each block as the average of all similarity values within it:
    \begin{equation}
        \mathcal{C}_i = \mathcal{C}(a_i, b_i, x_i, y_i) = \frac{\sum_{m=a_i}^{x_i}\sum_{n=b_i}^{y_i} \mathcal{M}_{mn}}{(x_i-a_i)(y_i-b_i)}.
    \label{eq:step}
    \end{equation}
    Eq.~\ref{eq:step} measures the step consistency between two videos within the period covered by the $i$-th block. A higher $\mathcal{C}_i$ indicates that these two videos are more step-wise consistent.
    Given a relational matrix $\mathcal{M}$, our objective is to find a $K$-block partition of the matrix that maximizes the cumulative sum of the step consistency scores. This way, the partition is regarded as the optimal step segmentation of these two videos. Specifically, the objective can be expressed as:
    \begin{equation}
        \underset{\mathcal{B}}{\text{maximize}}\ {\sum_{i=1}^K \mathcal{C}(a_i,b_i,x_i,y_i)}.
        \label{eq:max_goal}
    \end{equation}
    Eq.~\ref{eq:max_goal} involves many variables to be optimized, making it difficult to be solved directly. Fortunately, resorting to dynamic programming allows us to address it efficiently. Specifically, we devise a three-dimensional dynamic step mining algorithm equipped with a dynamic table $D_\text{csm}$ to find an optimal solution for Eq.~\ref{eq:max_goal}. $D_{\mathrm{csm}}(i, j, k)$ represents the sum of $k$ consistency scores composed of the first $i$ frames $f_{1:i}^1$  of video-1 and the first $j$ frames $f_{1:j}^2$ of video-2.

    For the boundary condition $k=1$, we can easily get the dynamic table's value by:
    \begin{equation}
        D_{\mathrm{csm}}(i, j, 1) = \mathcal{C}(1, 1, i, j).
    \end{equation}
    The cumulative sum of consistency scores at step $k$ can be calculated by adding the consistency score in $k_{\mathrm{th}}$ step to the cumulative sum of $\mathcal{C}$ at step $k-1$. Therefore, we can calculate $D_{\mathrm{csm}}(\cdot, \cdot, k)$ from $D_{\mathrm{csm}}(\cdot, \cdot, k-1)$, as illustrated in Fig.~\ref{fig:csm_method}(b). The update function of $D_{\mathrm{csm}}(i, j, k)$ is:
    
    \begin{equation}
        D_{\mathrm{csm}}(i, j, k) = \max_{a\leq i, b\leq j} \left[ D_{\mathrm{csm}}(a, b, k-1) + \mathcal{C}(a, b, i, j)\right]
    \end{equation}
    The value $D_{\mathrm{csm}}(T, T, K)$ means the maximal consistency score of partitioning video-1 and video-2 into $K$ steps.

\subsubsection{Step segmentation backtracing}
    Upon obtaining the optimal score, we need to retrace the alignment decisions to achieve a coherent step segmentation. Therefore, during the forward phase, we also need to record the selected indices of each update in the dynamic index table $D_{\mathrm{id}}$:
    \begin{equation}
        D_{\mathrm{id}}(i, j, k) = \argmax_{a, b;\ a\leq i, b\leq j} \left[ D_{\mathrm{csm}}(a, b, k-1) + \mathcal{C}(a, b, i, j) \right]
    \end{equation}
    Then, starting from the end $D_{\mathrm{id}}(T, T, K)$, we trace back to get each step's boundary, which is presented in Algorithm~\ref{alg:backtrace}.

    \begin{algorithm}[t]
        \renewcommand{\algorithmicrequire}{\textbf{Input:}}
        \renewcommand{\algorithmicensure}{\textbf{Output:}}
        \caption{Step segmentation backtracing}
        \label{alg:backtrace}
        \begin{algorithmic}[1]
            \REQUIRE The dynamic index table $D_{\mathrm{id}}$; The required step number $K$.
              \ENSURE Two videos' step boundaries $B_1$ and $B_2$.
              \STATE Initialization: $B_{1}\leftarrow \{\}$, $B_{2}\leftarrow \{\}$, $k\leftarrow K$, $i\leftarrow T$, $j\leftarrow T$
              \REPEAT
              \STATE $a,b\leftarrow D_{\mathrm{id}}[i, j, k]$
              \STATE append $a$ to $B_1$
              \STATE append $b$ to $B_2$
              \STATE $k\leftarrow k-1$;\ $i\leftarrow a$;\ $j\leftarrow b$
              \UNTIL x==0
        \end{algorithmic}  
    \end{algorithm}

    Through backtracing, we can get two videos' step boundaries. Within each step, we randomly sample one frame-level feature to act as the representation of the step. Then we can get the step-level features $S_1=\{s_1^1, s_2^1, \dots, s_K^1\}$ and $S_2=\{s_1^2, s_2^2, \dots, s_K^2\}$, which is used in the following frame-to-step alignment stage. 

\subsection{Frame-to-Step Alignment}
    Based on the collaboratively mined steps of paired videos, we further propose a frame-to-step alignment (FSA) module. It calculates the probability of aligning the step-level features of one video with the frame-level features of another. A larger alignment probability indicates they are more likely to be step-wise consistent. The motivation behind this \textit{cross-verification} design is: \textit{if two videos are step-level consistent, then we can achieve a good alignment between the frames of video-1 and the steps of video-2, as video-2's step is extracted under the guidance of video-1.} Experiments in Sec.~\ref{sec:vis_align} also prove this argument.

    Given the frame-level features $F=\{f_1, \dots f_T\}$ of one video and the step-level features $S=\{s_1,\dots s_K\}$ of another video, the alignment between them is a dense-to-sparse ($T>K$) mapping that includes many possibilities. We denote one possible alignment as $\pi=(\pi_1, \pi_2,\dots, \pi_{T})$, which represents the frame-to-step assignments. The probability of alignment $\pi$ is calculated as:
    \begin{equation}
        p(\pi | F)=\prod_{t=1}^T p_{\pi_{t}}^t,\ \pi_t\in \{1,2,\dots, K\}
    \end{equation}
    where $p_{\pi_t}^t$ means the probability of assigning frame $t$ to step $\pi_t$. Here $\pi_t$ is one of the $K$ steps. For a given step-level features $S$, we can calculate the cumulative probability of all possible alignments with:
    \begin{equation}
        P(S|F)=\sum_{\pi\in \Omega(F,S)} p(\pi|F)
    \end{equation}
    where $\Omega(F,S)$ is the set of all possible frame-to-step alignments given $F$ and $S$. It is hard to compute all possible alignments, but we can also resort to dynamic programming to make the computation tractable. We first compute the frame-to-step probability matrix $\mathcal{W}=\mathrm{Softmax}\left[(F\cdot S^\mathrm{T}) / \sqrt{d}\right]$. From $\mathcal{W}$, we can get the frame-to-step assignment probability $p_{L_k}^t=\mathcal{W}_{tk}$. Then, we design a two-dimensional dynamic table $D_{\mathrm{fsa}}$ with shape $T\times K$ to record the alignment probabilities. $D_{\mathrm{fsa}}(t,k)$ means the probability of aligning frames $F_{1:t}$ to steps $S_{1:k}$.

    First, we define the boundary conditions: $D_{\mathrm{fsa}}(t, 1) = 1$. Then, we need to define the update function for $D_{\mathrm{fsa}}(t, k)$. At each timestamp $t$, we have two choices. If frame $t$ does not switch steps, then the probability comes from $D_{\mathrm{fsa}}(t-1, k)$. If frame $t$ switch step compared with the previous step, then the probability comes from $D_{\mathrm{fsa}}(t-1, k-1)$. Therefore, we can update the dynamic table with Eq.~\ref{eq:csm_update}:
    \begin{equation}
        D_{\mathrm{fsa}}(t,k)=\mathcal{W}_{tk}\left[ D_{\mathrm{fsa}}(t-1,k)+D_{\mathrm{fsa}}(t-1,k-1) \right]
        \label{eq:csm_update}
    \end{equation}
    Our goal is to get the probability of aligning frame features $F$ to step features $S$, which is:
    \begin{equation}
        p(S|F) = D_{\mathrm{fsa}}(T, K)
    \end{equation}
    where $D_{\mathrm{fsa}}(T, K)$ represents the complete frame-to-step alignment probability.

    We use the negative log-likelihood value as the procedure correlation distance. For two videos' frame-level features $F_1,F_2$ and step-level features $S_1,S_2$, we compute the alignment scores from two directions to maintain symmetry. The final distance $d_{\mathrm{align}}$ is:
    \begin{equation}
        d_{\mathrm{align}} = -\frac{1}{2}\left[\log P(S_2|F_1)+\log P(S_1|F_2)\right]
    \end{equation}
    
    Moreover, we design variants of our frame-to-step module for handling repetitive steps and background frames, which is detailed in the supplementary materials.
    
\subsection{Optimization}
    During training, we adopt three loss functions for optimization. First, from the CSM, we can get the cumulative sum of consistency score: $D_{\mathrm{csm}}(T, T, K)$. We design a step-enhancing loss $\mathcal{L}_{\mathrm{step}}$ to maximize the consistency scores of positive pairs to enhance the frame similarity within the same step. It can be formulated as: 
    \begin{equation}
        \mathcal{L}_{\mathrm{step}}=-D_{\mathrm{csm}}(T, T, K)
    \end{equation}
    Second, from the FSA, we can get the procedure correlation distance $d_{\mathrm{align}}$ between two videos. We design a aligning loss $\mathcal{L}_{\mathrm{align}}$ to minimize the $d_{\mathrm{align}}$ of positive pairs:
    \begin{equation}
        \mathcal{L}_{\mathrm{align}}=d_{\mathrm{align}}
    \end{equation}
    
    The third loss $\mathcal{L}_{\mathrm{task}}$ changes with the tasks. For sequence verification, we follow \cite{qian2022svip} to adopt procedure classification as an auxiliary task and its loss is:
    \begin{equation}
        \mathcal{L}_{\mathrm{task}}=\mathrm{Cross\text{-}Entropy}\left(pred, Y\right)
    \end{equation}
    where $pred$ and $Y$ are the procedure predictions and labels. For action quality assessment, we follow \cite{xu2022finediving} to optimize the mean squared error between the ground truth $y_X$ and predicted action score $\hat{y}_X$:
    \begin{equation}
        \mathcal{L}_{\mathrm{task}}=\lVert \hat{y}_X - y_X\rVert^2
    \end{equation}
    
    Hence, the overall loss $\mathcal{L}$ for optimization is:
    \begin{equation}
        \mathcal{L} = \mathcal{L}_{\mathrm{task}} +  \mathcal{L}_{\mathrm{step}} + \mathcal{L}_{\mathrm{align}}
    \end{equation}

\section{Experiments}
\subsection{Implementation details}
    We implement our method on two instructional video tasks, including sequence verification and action quality assessment. For sequence verification, our implementation adheres to~\cite{qian2022svip} for a fair comparison. Besides ResNet-50, we additionally utilize X3D-m pretrained on Kinetics-400~\cite{kay2017kinetics} as our backbone for experiments. For action quality assessment, our implementation sticks to the method described in~\cite{xu2022finediving} for a fair comparison. For all tasks, we trained our model on 2 NVIDIA TITAN RTX GPUs with batch size 8. More details can be found in supplementary materials.

\subsection{Sequence Verification}
\subsubsection{Goal}
    Sequence verification (SV) aims to verify whether two instructional videos have identical procedures. Two videos executing the same steps in the same order form a positive pair, otherwise negative. The method should give a verification distance between each video pair and give the prediction by thresholding the distance. We conduct experiments on three sequence verification datasets (CSV, Diving-SV, and COIN-SV) proposed by \cite{qian2022svip}. We adopt Area Under ROC Curve (AUC) for evaluation. A higher AUC indicates better performance.

\subsubsection{Competitors}
    We compare our method with various approaches, including (1) video methods: TRN~\cite{zhou2018temporal}, TSM~\cite{lin2019tsm}, Video-swin~\cite{liu2022video}, CAT~\cite{qian2022svip}; (2) sequence alignment methods: OTAM~\cite{cao2020few}, TAP~\cite{pan2021temporal}, Drop-DTW~\cite{dvornik2021drop}; (3) visual-language methods: CLIP+TE+MLP~\cite{radford2021learning}, WeakSVR~\cite{dong2023weakly}. Visual-language methods are pretrained on CLIP~\cite{radford2021learning}, while other methods are pretrained on Kinetics-400 (K-400)~\cite{kay2017kinetics}. For video and visual-language methods, we follow~\cite{dong2023weakly}'s setting to calculate the normalized L2 distance between two videos' representations as their verification distance. For sequence alignment methods, we use their alignment distance $d_{\mathrm{align}}$ as the verification distance, which is the same as our method.

    \begin{table}[H]
        \centering
        \begin{adjustbox}{max width=\linewidth}
        \begin{tabular}{c|c|c|c|c|c}
            \hline
    	\multirow{2}*{Method} & \multirow{2}*{Pretrain} & \multirow{2}*{\makecell{Text \\ Anno.}} & \multicolumn{3}{c}{AUC} \\
        \cline{4-6}
         & & & CSV & Diving-SV & COIN-SV \\
        \hline
        TRN & \multirow{4}*{K-400} & \scalebox{0.75}{\usym{2613}} & 80.32 & 80.69 & 57.19 \\
        Video-Swin & & \scalebox{0.75}{\usym{2613}} & 54.06 & 73.10 & 43.70 \\
        CAT & & \scalebox{0.75}{\usym{2613}} & 83.02 & 83.11 & 51.13 \\
        \cdashline{1-6}[1pt/1pt]
        OTAM & \multirow{3}*{K-400} & \scalebox{0.75}{\usym{2613}} & 69.03 & 77.86 & 50.55 \\
        TAP & & \scalebox{0.75}{\usym{2613}} & 73.29 & 75.47 & 47.45 \\
        Drop-DTW & & \scalebox{0.75}{\usym{2613}} & 84.86 & 74.12 & 53.33 \\
        \cdashline{1-6}[1pt/1pt]
        CLIP+TE+MLP & \multirow{2}*{CLIP} & \checkmark & 79.38 & 83.48 & 48.50 \\
        WeakSVR & & \checkmark & 86.92 & 86.09 & \textbf{59.57} \\
        \cdashline{1-6}[1pt/1pt]
        CPA+R50 (ours) & \multirow{2}*{K-400} & \scalebox{0.75}{\usym{2613}} & \textbf{88.14} & 84.29 & 57.57 \\
        CPA+X3D (ours) & & \scalebox{0.75}{\usym{2613}} & 86.06          & \textbf{88.11} & 57.55 \\
        \hline
        \end{tabular}
        \end{adjustbox}
        \caption{Results of sequence verification.}
        \label{tab:perform}
    \end{table}
    
\subsubsection{Results}
    The results are presented in Tab.~\ref{tab:perform}. For CSV, our method achieves the best performance (88.14\%). For Diving-SV, our method gets the best performance among methods with a 2D backbone. Furthermore, applying the 3D backbone X3D further boosts our performance to the new state-of-the-art result (88.11\%). For COIN-SV, our approach achieves the best among visual-based methods. It is worth noticing that our method remains competitive even among visual-language methods, despite these methods being equipped with extra text narrations on procedures.

\subsection{Action Quality Assessment}
\subsubsection{Goal}
    In action quality assessment (AQA), we adopt \cite{xu2022finediving}'s setting, where an exemplar video and its score are given. Then for a query video with the same procedure, the method should predict its score based on the exemplar. We conduct experiments on the FineDiving dataset. We use Spearman's rank correlation ($\rho$) and relative $\ell 2$-distance (R-$\ell 2$) for evaluation. A Higher $\rho$ and lower R-$\ell 2$ indicate better performance. Average Intersection over Union (AIoU) is adopted to evaluate the procedure segmentation.

\subsubsection{Competitors}
    We compare our approach with various advanced AQA methods, including (1) non-procedure methods: USDL, MUSDL~\cite{tang2020uncertainty}, CoRe~\cite{yu2021group}, Lian et al.~\cite{lian2023improving}; (2) procedure-aware method: TSA~\cite{xu2022finediving}, PECoP~\cite{dadashzadeh2023pecop}. Vanilla TSA's procedure segmentation module is supervised by step-level annotations. Here we replace TSA's procedure segmentation module with our CPA \emph{to achieve segmentation without the help of step-level annotations.}
    
    \begin{table}[H]
        \centering
        \begin{adjustbox}{max width=\linewidth}
        \begin{tabular}{c|c|cc|c|c}
            \hline
            \multirow{2}*{Method} & \multirow{2}*{\makecell{Step \\ Anno.}} & \multicolumn{2}{c|}{AIoU@} & \multirow{2}*{$\rho$} & \multirow{2}*{R-$\ell 2\ (\times 100)$} \\
            \cline{3-4}
                    &       &  0.5  & 0.75  &           & \\
            \hline
             USDL   & \scalebox{0.75}{\usym{2613}} & / & / & 0.8913 & 0.3822 \\
             MUSDL  & \scalebox{0.75}{\usym{2613}} & / & / & 0.8978 & 0.3704 \\
             CoRe   & \scalebox{0.75}{\usym{2613}} & / & / & 0.9061 & 0.3615 \\
             Lian et al. & \scalebox{0.75}{\usym{2613}} & / & / & 0.9222 & 0.3304 \\
             PECoP & \checkmark & - & - & 0.9315 & - \\
             TSA    & \checkmark &  82.51 & 34.31 & 0.9203 & 0.3420 \\
             TSA (our impl.) & \checkmark & 93.23 & \textbf{53.39} & 0.9302 & 0.3154 \\
             \cdashline{1-6}[1pt/1pt]
             CPA+TSA (ours) & \scalebox{0.75}{\usym{2613}} & \textbf{94.28} & 21.14 & \textbf{0.9364} & \textbf{0.2909} \\
             \hline
        \end{tabular}
        \end{adjustbox}
        \caption{Results of action quality assessment on FineDiving. / indicates ``without procedure segmentation''.}
        \label{tab:frame-step}
        \label{tab:aqa}
    \end{table}
    
\subsubsection{Results}
    The results are presented in Tab.~\ref{tab:aqa}. Our method achieves new state-of-the-art results. Note that our method outperforms TSA without step-level annotations. The reason is that TSA strictly divides the procedure into three steps: \emph{take-off}, \emph{flight}, and \emph{entry}, which might not be the most suitable division for assessment. In contrast, our CPA can flexibly adjust the procedure segmentation through optimization to get better assessments. Furthermore, by observing the AIoU, our CPA can get satisfying coarse-grain procedure segmentation (AIoU@0.5=94.28). Note that we uniformly divide CPA's segments into three steps for calculating AIoU. Finally, we can find that procedure-aware methods generally outperform non-procedure methods, which further emphasizes the importance of procedural knowledge for accurate predictions on instructional videos.

\begin{figure*}
    \centering
    \includegraphics[width=0.9\textwidth]{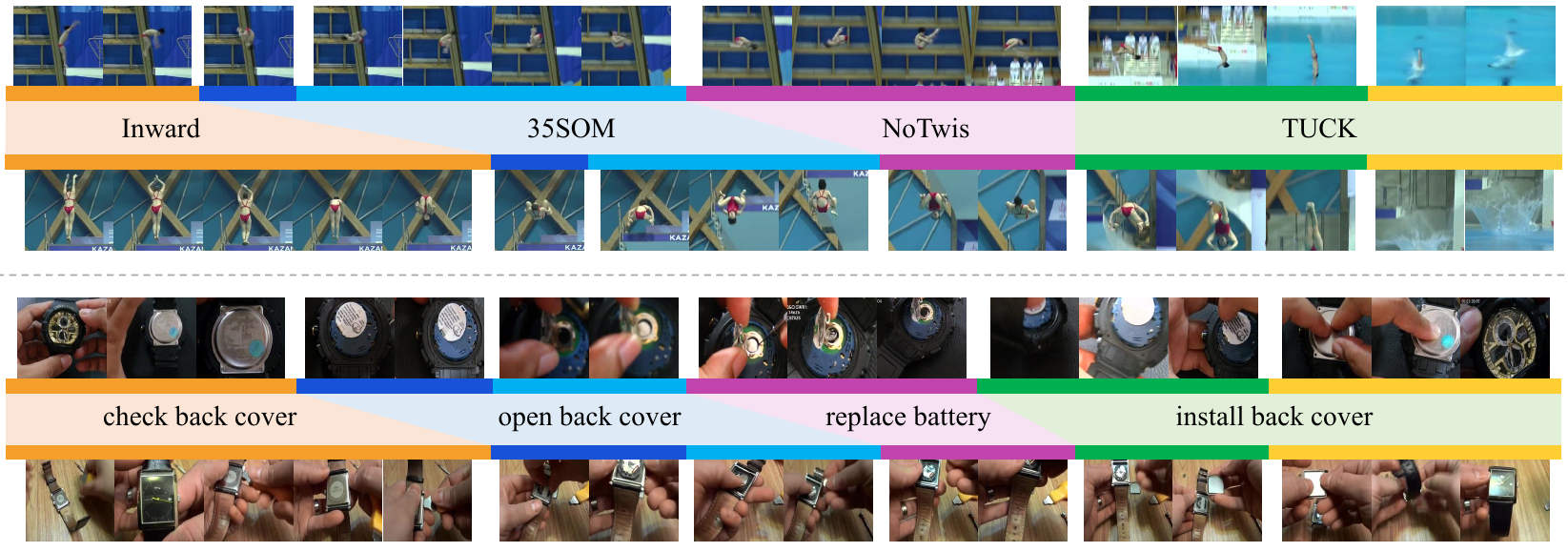}
    \caption{Visualization of collaborative step segmentation with $T=16$ and $K=6$.}
    \label{fig:segvis_clustering}
\end{figure*}

\subsection{In-depth Analysis}
\subsubsection{\textbf{Ablations on main components}}
    We conduct an ablation analysis of the proposed modules on SV and AQA. Results summarized in Tab.~\ref{tab:compare} demonstrate the effectiveness of each module. Note that using FSA individually means we just uniformly sample $K$ frame features as the step-level features. In the first row of SV, we use the video-level features to calculate the verification distance. In the first two rows of AQA, we uniformly divide the video into $K$ segments. According to the results, using FSA enable the method to learn more distinctive frame-level features and improve performance on both tasks. Moreover, introducing CSM contributes significantly to improvements, highlighting the importance of step information in this task and the effectiveness of our CSM on step mining. 
    \begin{table}[H]
        \centering
        \begin{adjustbox}{max width=\linewidth}
        \begin{tabular}{cc|c|cc}
            \hline
            \multirow{2}*{CSM} & \multirow{2}*{FSA} & SV on CSV & \multicolumn{2}{c}{AQA on FineDiving} \\
            \cline{3-5}
                       &            & AUC & $\rho$ & R-$\ell 2$ \\
            \hline
                       &            & 81.65 & 0.9221 & 0.3456 \\
                       & \checkmark & 86.23 & 0.9275 & 0.3288 \\ 
            \checkmark & \checkmark & \textbf{88.14} & \textbf{0.9364} & \textbf{0.2909} \\
            \hline
        \end{tabular}
        \end{adjustbox}
        \caption{Ablation study of the proposed modules. CSM: Collaborative Step Mining; FSA: Frame-to-Step Alignment.}
        \label{tab:compare}
    \end{table}

\subsubsection{Analysis on step number}
    In real-world videos, the definition of a step is flexible and ambiguous, where several adjacent steps can be reorganized as one step at a coarser level. Our CSM can provide step-mining results across different step granularity. Fig.~\ref{fig:multi_step} illustrates the results of CSM under different step numbers $K$, where each block signifies a step. Our method degrades to frame-to-frame comparison when $K=T$. We further conduct sensitivity analysis on SV and AQA by adjusting $K$. From Tab.~\ref{tab:step_num}, we can observe that the performances first rise and then drop as the step number increases. The performances peak at $K=13$ for CSV and $K=6$ for FineDiving, whose trend corresponds with the maximum step number in CSV (16 steps) and FineDiving (5 steps). Therefore, choosing the proper step number depending on the dataset can lead to better performance.
    
    \begin{table}[t]
        \centering
        \begin{adjustbox}{max width=0.9\linewidth}
        \begin{tabular}{p{1cm}<{\centering}|p{1cm}<{\centering}p{1cm}<{\centering}p{1cm}<{\centering}p{1cm}<{\centering}p{1cm}<{\centering}}
            \Xhline{0.8pt}
            \multicolumn{6}{c}{Sequence Verification} \\
            \hline
            \multirow{2}*{Metric} & \multicolumn{5}{c}{Step number $K$} \\
            \cline{2-6}
             & 11 & 12 & 13 & 14 & 15 \\
            \hline
            AUC & 84.91 & 87.00 & \textbf{88.14}  & 87.94 & 84.67\\
            \Xhline{0.8pt}
        \end{tabular}
        \end{adjustbox}
        \begin{adjustbox}{max width=0.9\linewidth}
        \begin{tabular}{p{1cm}<{\centering}|p{1cm}<{\centering}p{1cm}<{\centering}p{1cm}<{\centering}p{1cm}<{\centering}p{1cm}<{\centering}}
            \multicolumn{6}{c}{Action Quality Assessment} \\
            \hline
            \multirow{2}*{Metric} & \multicolumn{5}{c}{Step number $K$} \\
            \cline{2-6}
             & 3 & 4 & 5 & 6 & 7 \\
            \hline
            $\rho$      & 0.9302 & 0.9309 & 0.9316 & \textbf{0.9364} & 0.9340 \\
            R-$\ell 2$  & 0.3098 & 0.3202 & 0.3119 & \textbf{0.2909} & 0.2943 \\
            \Xhline{0.8pt}
        \end{tabular}
        \end{adjustbox}
        \caption{Sensitivity analysis on step number $K$.}
        \label{tab:step_num}
    \end{table}
    
\begin{figure}[t]
    \centering
    \includegraphics[width=\linewidth]{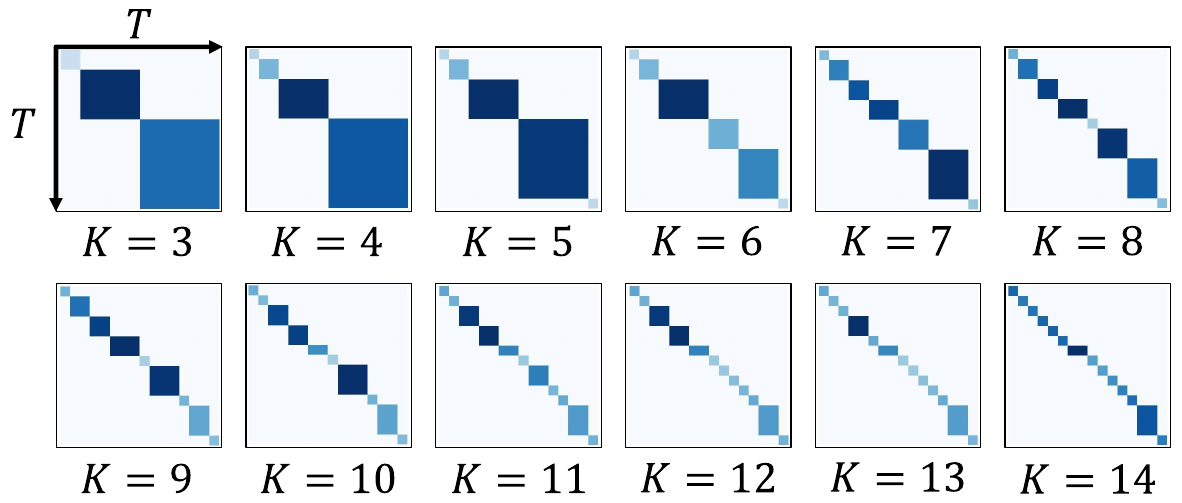}
    \caption{The block-diagonal structures representing the multi-grained step segmentation, experimented on CSV.}
    \label{fig:multi_step}
\end{figure}

\subsubsection{Collaborative step mining visualization}
    \label{sec:csm_vis}
    We visualize step segmentation produced by CSM in Fig.~\ref{fig:segvis_clustering}(a) on ``diving'' and ``changing watch battery''. We set $T=16$, $K=6$, and manually name the text descriptions of steps for better visualization. By observation, some steps (e.g., ``35som'' and ``install back cover'') are further divided into more steps in our six-step segmentation. The reason is that step segmentation can sometimes be flexible and multi-grained. CSM can well identify consistent steps, thereby providing dependable guidance for subsequent frame-to-step alignment. 

\subsubsection{Frame-to-step alignment visualization}
\label{sec:vis_align}
\begin{figure}[t]
      \centering
      \includegraphics[width=\linewidth]{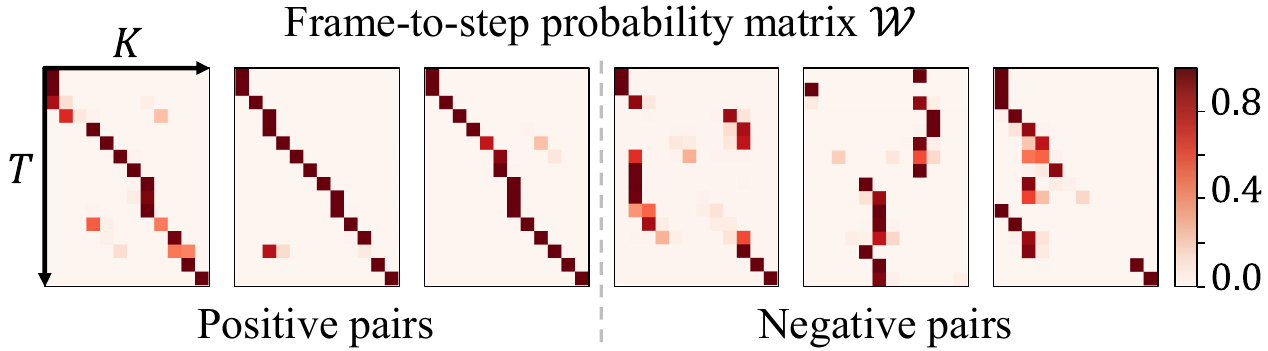}
      \caption{The frame-to-step aligning probability on CSV, with three positive and negative pairs. The vertical and horizontal axes represent $T$ frames and $K$ steps respectively.}
      \label{fig:fsa_vis}
    \end{figure}

    As depicted in Fig.~\ref{fig:fsa_vis}, we show the frame-to-step assignment probabilities for both positive and negative pairs on SV. By observation, the positive pairs can form continuous probability paths from the top-left corner to the bottom-right corner, while negative pairs will exhibit obvious discontinuities. Consequently, paired videos adhering to a chronological procedural alignment can achieve minimal distance using our proposed method, which validates our motivation for the cross-verification design in frame-to-step alignment.

\begin{figure}[t]
    \centering
    \includegraphics[width=0.95\linewidth]{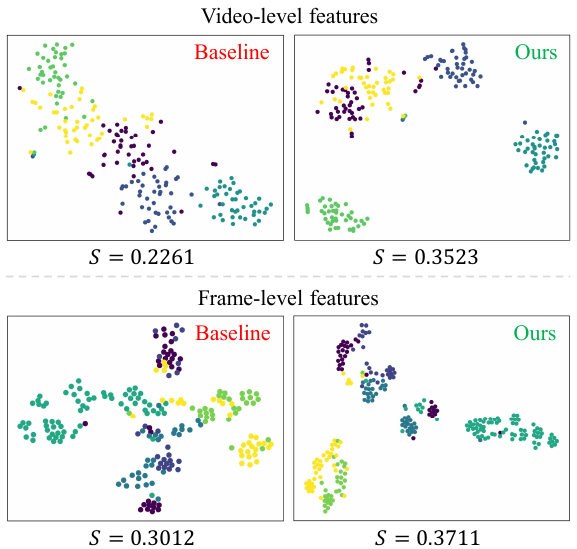}
    \caption{The t-SNE visualization of multi-level features and their respective Silhouette scores $S$. top: the video-level features with colors indicating video classes. Bottom: the frame-level features with colors indicating step classes.}
    \label{fig:multi_step}
\end{figure}

\subsubsection{Multi-level feature visualization}
\label{sec:step_cluster}
    We demonstrate our method's feature learning ability by using t-SNE to visualize both the video-level and frame-level features trained on CSV, which is shown in Fig.~\ref{fig:segvis_clustering}(b). For video-level features, the color represents its video-level class. For frame-level features, the color represents its step, which is manually annotated for visualization. Besides qualitative visualization, we also adopt the Silhouette score $S$ to quantify the clustering effect. A higher Silhouette score indicates a better clustering outcome. From Fig.~\ref{fig:segvis_clustering}(b), compared with baseline, our method improves the feature clustering effects on both the video-level features (0.2261 $\to$ 0.3523) and the frame-level features (0.3012 $\to$ 0.3711). This result indicates that our method can learn more distinctive multi-level features, which is beneficial for instructional video analysis.

\section{Conclusion}
    In this paper, we propose a weakly supervised framework for procedure-aware correlation learning on instructional videos, named the Collaborative Procedure Alignment (CPA). Under this framework, we first design the collaborative step mining (CSM) module to simultaneously produce step segmentation for paired videos and get the representative step-level features. Furthermore, we propose the frame-to-step alignment (FSA) module to calculate the correlation distance between videos. Extensive and in-depth experiments on two instructional video tasks showcase the superiority of our framework, demonstrating its capacity to deliver more explainable and accurate understandings on complex instructional videos.

\section{Acknowledgments}
    The paper is supported in part by the National Natural Science Foundation of China (No. 62325109, U21B2013, 61971277) and the Lenovo Academic Collaboration Project.

\bibliography{aaai24}

\end{document}